\documentclass{llncs}
\usepackage{amsmath}
\usepackage{amssymb}
\usepackage{graphicx}
\usepackage{url}

\title{Electronic Geometry Textbook: A Geometric Textbook Knowledge Management System\thanks{The final publication of this paper is available at www.springerlink.com.}}
\author{Xiaoyu Chen}
\institute{LMIB -- SKLSDE -- School of Mathematics and Systems
Science,\\ Beihang University, Beijing 100191, China}
\date{}

\begin{document}
\bibliographystyle{plain}
\maketitle

\begin{abstract}
Electronic Geometry Textbook is a knowledge management system that
manages geometric textbook knowledge to enable users to construct
and share dynamic geometry textbooks interactively and efficiently.
Based on a knowledge base organizing and storing the knowledge
represented in specific languages, the system implements interfaces
for maintaining the data representing that knowledge as well as
relations among those data, for automatically generating readable
documents for viewing or printing, and for automatically discovering
the relations among knowledge data. An interface has been developed
for users to create geometry textbooks with automatic checking, in
real time, of the consistency of the structure of each resulting
textbook. By integrating an external geometric theorem prover and an
external dynamic geometry software package, the system offers the
facilities for automatically proving theorems and generating dynamic
figures in the created textbooks. This paper provides a
comprehensive account of the current version of Electronic Geometry
Textbook.
\end{abstract}

\section{Introduction}

\subsection{Motivation}

When we speak about managing knowledge, we may start by thinking about
textbooks, where knowledge is organized systematically and presented
hierarchically according to its internal logical relations. Since textbooks provide
a well-arranged structure of domain knowledge, they play an
important role in education and research; they record knowledge and impart
it to new learners. The Electronic Geometry Textbook (EGT) is a
knowledge management system for geometric knowledge, built so that users may
construct and publish dynamic geometry textbooks interactively and
efficiently. The objective of our textbook project is to explore the
approaches to managing knowledge by integrating available software
tools and providing a system that assists human authors to create
dynamic, interactive, and machine-processable textbooks (instead of
the traditional static textbooks). EGT is motivated by the following
three considerations:

(1) Textbooks are a standard form for the storage, organization, and
presentation of systematic domain knowledge. For different
pedagogical purposes, the same knowledge may be adopted by different
textbooks as a part of the theories involved. In order to share and
reuse sophisticated knowledge, we need to build up a standard
knowledge base that stores and organizes data describing
the textbook knowledge. Authors can contribute knowledge (as data encoded in some
knowledge representation format) to the knowledge base and
\emph{construct} textbooks by reusing pieces of knowledge already in
the knowledge base as constituents in a new textbook.

EGT offers such an environment that maintains
(i.e., creates, removes, modifies, and queries, etc.) and shares
knowledge data with an appropriate granularity, constructs textbooks
by interactively arranging the knowledge data selected and retrieved
from the knowledge base, and automatically generates styled
documents for browsing and printing the textbooks produced.

(2) When creating a textbook for learners, one needs to determine an
appropriate narrative structure so as to arrange the contents
involved in the textbook. Although one can make one's own decision
as to what knowledge is to be chosen, there are common practices and
implicit conventions in a scientific community as to how knowledge
should be organized, formulated, and presented. For example, it is
commonly accepted that a proved proposition is a lemma only if it is
used in the proof of a theorem and a corollary is a true proposition
that follows from a theorem. The domain knowledge presented in a
textbook should be structured systematically, hierarchically, and
logically, i.e., from the simplest to the most complicated and from
the basic to the advanced. For example, the definition for each
concept in a statement (such as a theorem, exercise, or example
statement) should have been given before the statement. In order to
produce such a sound and usable textbook, we need to be given
feedback if the narrative structure disobeys the conventional rules
during the process of construction.

EGT offers such a facility that assists a user
to automatically check, in real time, whether the constructed textbook has a
satisfiable and reasonable narrative structure.

(3) In recent years, many creative methods have been proposed for
automated geometry theorem proving, such as algebraic approaches
(the most powerful, although just decision methods) which convert
the problem of geometrical reasoning to that of solving algebraic
systems, coordinate-free approaches which convert the problem to the
counterpart of algebraic calculation with respect to some geometric
quantities, and traditional AI approaches. \cite{reasoning} Many
geometry software tools have implemented these approaches and
provided the functionality of automated reasoning, such as GEOTHER
\cite{geother}, Geometry Expert \cite{gex}, and GCLC \cite{gclc},
and of interactive proving, such as GeoProof \cite{geoproof}. These
tools not only help geometry researchers to discover new and more
valuable and complex theorems, but also support geometry education.
Geometry textbooks include many interesting and complicated theorems
whose proofs are given and checked by the authors. As the
traditional formal logical methods do not work very efficiently in
automated geometry deduction, the techniques for automated geometric
proof checking have not been well developed. However, it is still
helpful to make use of the automated theorem provers to assist an
author to determine whether a proposition written in a textbook is
logically true in order to ensure the correctness of the textbook.

In addition, geometry deals with graphical objects abstracted from
the real visual world. Intuitive figures are indispensable
constituents of textbooks. With the help of a computer, one can draw
high-resolution and accurate figures interactively by using a mouse
and following the instructions provided during the construction. For
instance, after selecting two points and an instruction ``make the
mid-point of two points" by using a mouse, the mid-point will be
constructed and displayed in the figure. An even more important
enhancement resulting from this interactive facility is that the
steps of constructing a figure can be recorded and redone quickly.
For example, as a result one drags a free point from one place to
another and the figure will be updated immediately. One can explore
a figure and experience what happens when components are moved. This
dynamic feature makes geometry more vivid. Dynamic geometry software
has been developed to implement these features and applied in
geometry education and research, such as Cabri \cite{cabri},
SketchPad \cite{sketchpad}, Cinderella \cite{cinderella}, and
GeoGebra \cite{geogebra}. It is useful to apply dynamic geometry
software to make the figures in textbooks more intuitive and
explorable than the traditional static ones.

EGT offers interfaces to knowledge data and to selected external
geometry software packages for proving theorems and generating
dynamic figures automatically.

\subsection{Originality}

The idea of designing and developing such an integrated software
system in the form of a textbook for systematic and interactive
management of geometric knowledge originates from Dongming Wang
\cite{geotextbook} who has been working on automated geometric
reasoning for the last two decades. The author has been stimulated
to elaborate the idea and to undertake the implementation of a
system himself. We consider geometry a unique and rich subject of
mathematics that should be chosen for study in the context of
knowledge management. In such a study, the full power of computers
for symbolic, numeric, and graphical computing and data processing
may be used and our ideas may be effectively tested.

Several e-learning and intelligent tutoring systems for mathematics
have been proposed and developed, such as LeActiveMath
\cite{leactivemath}, ActiveMath \cite{activemath}, and MathDox
\cite{mathdox}. They offer facilities for generating courseware
which adapts to students, tutoring students interactively with diagnoses
of mistakes and adaptive feedback, analyzing and evaluating students' abilities,
etc. These systems are learner-centred and support the learner's
initiative. However, EGT is designed mainly to assist human
authors in constructing dynamic textbooks. The process is mostly
author-driven and manipulations of the textbook are allowed and may
lead to new, modified, or improved versions of a textbook. EGT's
innovations may be seen in the following three aspects:

\begin{enumerate}

\item EGT products can be viewed or printed as traditional textbooks (static documents) and also run as
dynamic software on a computer. Textbook knowledge is shared at an
appropriate granularity and textbooks can be constructed and
maintained interactively. For example, a textbook can be seen as an
arrangement of nodes that refer to the corresponding textbook
contents. One can perform a series of manipulations adding,
inserting, removing, modifying, and restructuring the nodes involved,
and meanwhile the generated documents for browsing and printing can
be updated automatically;

\item EGT can assist users to analyze the narrative structure
of the textbooks constructed and automatically find the parts
inconsistent with the conventional rules for writing textbooks, in
real time. We call this process \emph{consistency-checking} of the
structure of the textbook. For example, the definition of a
\texttt{median} of a triangle can be created only if the definition
of \texttt{midpoint} of a segment has already been introduced in the
textbook;

\item EGT integrates stand-alone geometry software packages for
automated theorem proving and dynamic figure generation. This
provides the constructed textbooks with dynamic features. For
example, the theorems in the textbooks can be automatically proved
by invoking geometric theorem provers, and the figures are
automatically constructed by applying dynamic geometry software.

\end{enumerate}

This paper describes the relevant design principles of EGT including
architectural issues, the structure of the geometric knowledge base,
knowledge representation, and the communication with available
geometry software packages. We present the main features of the
current version of EGT including maintaining geometric knowledge
data for constructing textbooks interactively, rendering the
textbooks in readable documents both in English and Chinese, proving
the theorems and drawing the dynamic figures automatically by
interfacing with the selected geometry software packages. While
plane Euclidean geometry is the target of our current investigation,
the ideas also apply to, or invite the attempt to apply them to,
other geometries.

\section{Design Principles of Electronic Geometry Textbook}

We describe the architecture of the system and present the main
design principles for a geometric textbook knowledge base and the
representation of knowledge. More details about the design
methodology have been discussed in \cite{dke}.

\subsection{Architecture and Communication}
\begin{figure}
\begin{center}
\includegraphics[width=10.5cm]{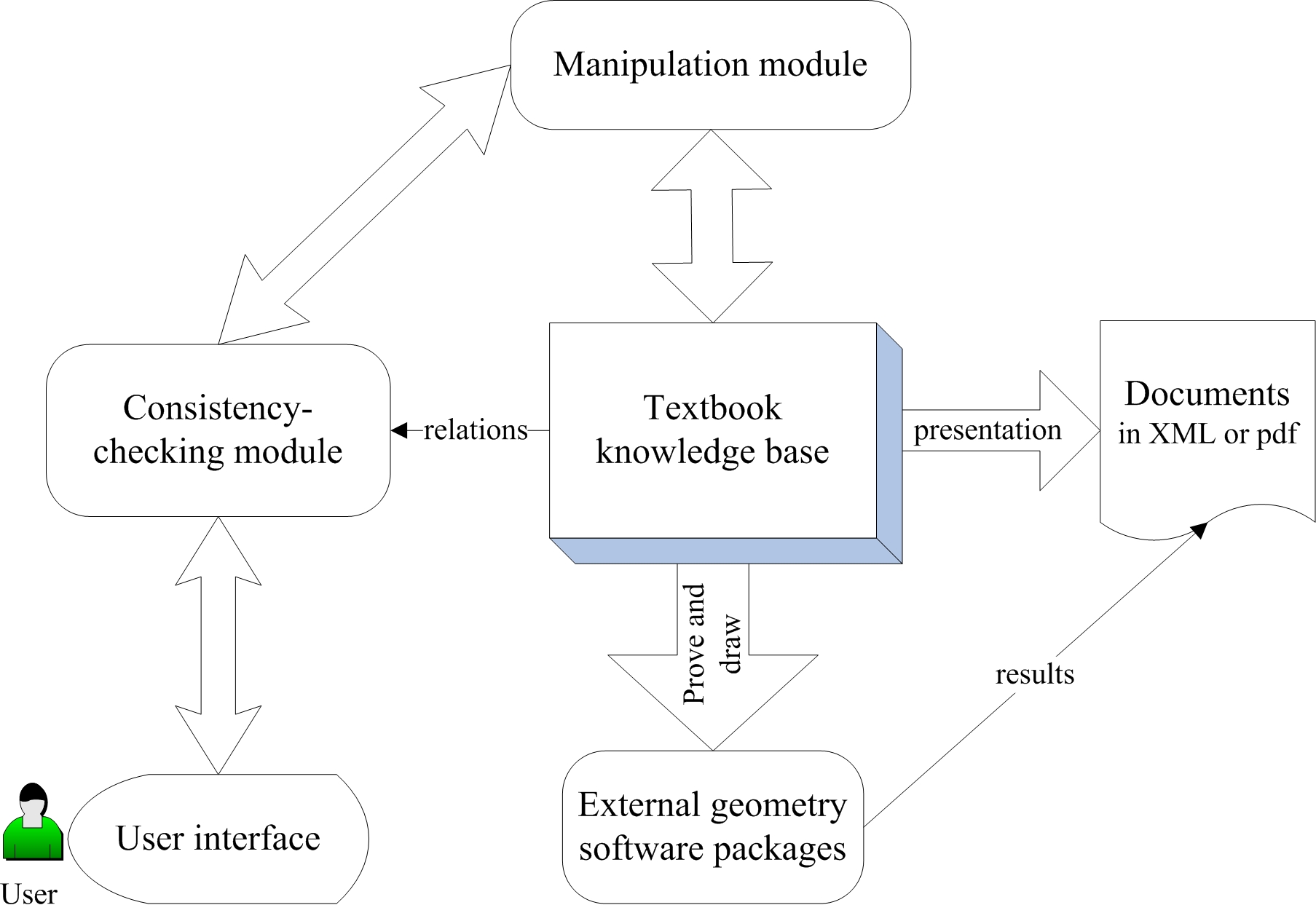}
\caption[1]{Architecture of the Electronic Geometry Textbook
system}\label{architecture}
\end{center}
\vspace{-0.5cm}
\end{figure}

Now we give a bird's eye view of how the system works and which
components carry out which tasks. In what follows, we refer to
Fig.~\ref{architecture} which gives an overview over the EGT
components and their communications. The \emph{textbook knowledge
base} is the kernel component of the system, storing and organizing
the shared knowledge data. Via the \emph{user interface}, users can
construct textbooks by invoking the \emph{manipulation module} to
perform the manipulations of creating new knowledge data, retrieving
needed knowledge data, and modifying knowledge data on the textbook
knowledge base. Meanwhile, the \emph{consistency-checking module}
will check the consistency of the constructed textbooks in real time
and provide feedback to the user interface. The textbooks
constructed can be presented in readable documents for rendering in
a browser or printing on paper. The theorems in the constructed
textbooks can be proved and dynamic figures can be drawn
automatically by interfacing with external geometry software
packages.

From the description of the system, we can conclude that the system
has as its foundation a textbook knowledge base. To
manipulate knowledge efficiently and appropriately, one
main task is creating a well-structured, manageable, and suitable
knowledge base.  On the other hand, to communicate knowledge,
formalizing and representing knowledge as data in a
processable way is the other main task.

\subsection{Design of Geometric Textbook Knowledge Base}

Generally speaking, the design of a knowledge base involves the
following two aspects.

\subsubsection{Knowledge Data Granularity.}

While creating a textbook, the author should identify the objects of
domain knowledge, categorize them, and rank them
according to their relationships. In order to support the
manipulations of constructing dynamic textbooks interactively, we
need to encapsulate knowledge data at an appropriate granularity. If
the granularity is too fine, the process of constructing textbooks
may be too complicated to manage. If the granularity is not fine
enough, the process may be trivial and not subject to manipulation.

We use the notion of a \emph{knowledge object} to represent the unit of textbook
knowledge which can be recognized, differentiated, understood, and
manipulated while constructing textbooks. For example, the
definition of a concept is a knowledge object which gives meaning to
the concept; a theorem is a knowledge object which is a true
proposition in the domain; a proof demonstrates that a proposition
is true; an exercise or example needs to be solved by applying some
knowledge. Working from the common or implicit conventions in
traditional textbooks, we classify the knowledge objects into the
following types: Concept (Definition), Axiom, Lemma, Theorem,
Corollary, Conjecture, Proof, Problem, Example, Exercise, Solution,
Algorithm, Introduction, and Remark. Although this classification
may be argued over and needs to be justified, what is essential in our
approach is to encapsulate knowledge data into certain knowledge
objects with the same structure. Within different types of knowledge
objects, certain data items are created to store knowledge data for
different applications on the objects. For example, the data stored
in a data item \emph{naturalRepresentation} is used for presentation,
the data stored in \emph{algebraicRepresentation} is used for
automatic proving by algebraic methods, the data stored in
\emph{diagramInstruction} is used for automatic dynamic figure
drawing, etc. One may refer to \cite{knowledgebase} for the details
of the  design of the structure within each type of knowledge object.

For a textbook, the index is an important component; it allows a
reader to see what is included and to navigate within the textbook. We use
\emph{category} to represent such a hierarchical structure so that a
\emph{category object} has a group of subcategories or knowledge
objects as its members. For example, each chapter in the textbook is
a category which usually has subcategories of sections, and each
section may include a group of knowledge objects. The categories
should usually be contributed by the authors using their
comprehensive understanding of the domain knowledge.

The textbook can be viewed as a linear arrangement of knowledge
objects and categories. The process of constructing textbooks can be
viewed as a series of manipulations (adding, inserting, removing,
modifying, and restructuring) of these knowledge objects and
categories.

\subsubsection{Relations.}

Geometric knowledge is accumulated step by step, e.g., by
introducing new concepts using already defined concepts, deriving
useful properties about new concepts, and proving or discovering
theorems relating old and new concepts. It does not lie flat but is
piled up with a certain intrinsic structure of hierarchy. Some
knowledge pieces serve as preliminaries for higher-level knowledge.
The conventional rules for writing textbooks depend on the
relationships of the knowledge involved. Therefore, the relations
among category objects and knowledge objects must be captured to
define the structure of the geometric textbook knowledge base, and
then to assist users to perform consistency-checking of the
structure of textbooks.

The relations we are interested in may involve the consideration of, and
abstractions from, pedagogical rules and textbook writing conventions.
We have identified 17 types of relations among knowledge objects and
category objects: Inclusion ($\rightarrow_{\rm include}$), Context
($\rightarrow_{\rm contextOf}$), Inheritance ($\rightarrow_{\rm
inherit}$), Derivation ($\rightarrow_{\rm deriveFrom}$), Implication
($\rightarrow_{\rm imply}$), Property ($\rightarrow_{\rm
hasProperty}$), Decision ($\rightarrow_{\rm decide}$), Justification
($\rightarrow_{\rm justify}$), Introduction ($\rightarrow_{\rm
introduce}$), Remark ($\rightarrow_{\rm remarkOn}$), Complication
($\rightarrow_{\rm complicate}$), Solution ($\rightarrow_{\rm
solve}$), Application ($\rightarrow_{\rm applyOn}$), Equality
($\leftrightarrow_{\rm equal}$), Exercise ($\rightarrow_{\rm
exerciseOf}$), Example ($\rightarrow_{\rm exampleOf}$), Association
($\leftrightarrow_{\rm associate}$).

The conventional rules for writing textbooks can be written with
these relations. For example, a relation $D\rightarrow_{\rm
contextOf}T$ (where $D$ is a definition and $T$ is a theorem) means
$D$ provides the context for $T$. The rule that $D$ should be
presented before $T$ in the textbook can be derived from the meaning
of the Context relation. Therefore, if $D$ is arranged after $T$
when a user constructs a textbook, then the structure of the
textbook is inconsistent and needs to be restructured. In
\cite{structure}, F.~Kamareddine et al.\ present an ontology and an
associated markup system for annotating mathematical documents so
that the graph of logical precedences (the conventional rules in our
context) of the annotated parts of text can be acquired and analyzed
automatically. However, we are concerned with not only how to
acquire these rules but also how to make use of them to decide
whether a textbook is constructed in an appropriate and soundly
presented structure, i.e., whether the structure of the textbook is
consistent. The inconsistencies found by the current system are
limited to those disobeying the rules derived from the existing
relations.

The geometric textbook knowledge base is then created to store textbook
knowledge data, with well-defined structures for the types of, and
relations between, the knowledge data stored.

\subsection{Knowledge Representation}
\label{language}

The knowledge data stored in the data items of knowledge objects
will be applied in different situations. One important application
is to communicate with external geometry software packages. It is
necessary to represent the geometric statements of the involved
knowledge objects in a formal language and to transform them
automatically into equivalent ones that the target geometry software
packages can identify and manipulate via specific interfaces. The
Intergeo project \cite{intergeo} is an ongoing European project, one
of whose objectives is to attack the barrier of lack of
interoperability by offering a common file format for specifying
dynamic figures. However, we have designed a geometry programming
language in which one can easily specify geometric statements of
definitions, theorems, axioms, and problems, etc.\ by using
customized concepts. We have also implemented automatic translation
of this language into the native languages of the geometry software
packages  targeted for communication. We present some examples using
this language here and describe how to process this language in
Section~\ref{communicationsection}.

\texttt{Simson's theorem} in English is ``The feet of the
perpendiculars from a point to the three sides of a triangle are
collinear if and only if the point lies on the circumcircle."

The formal representation of \texttt{Simson's theorem} is
``A\,:=\,point(); B\,:=\,point(); C\,:=\,point(); D\,:=\,point();
incident(D,\,circumcircle(triangle(A,\,B,\,C))) $\Leftrightarrow$
\hfil\break
collinear(foot(D,\,line(A,\,B)),\,foot(D,\,line(B,\,C)),\,foot(D,\,line(A,\,C)))."

The intersection point of two lines l and m is defined as\hfil\break
``intersection(l,\,m)\,$\triangleq$\,[A::Point where
incident(A,\,l)\,$\wedge$\,incident(A,\,m)]."

\section{Technical Realization of Electronic Geometry Textbook}

In this section, we present the technical details of implementing
the components of the Electronic Geometry Textbook system.

\subsection{Creation of the Geometric Textbook Knowledge Base}
\begin{figure}
\begin{center}
\includegraphics[width=11cm]{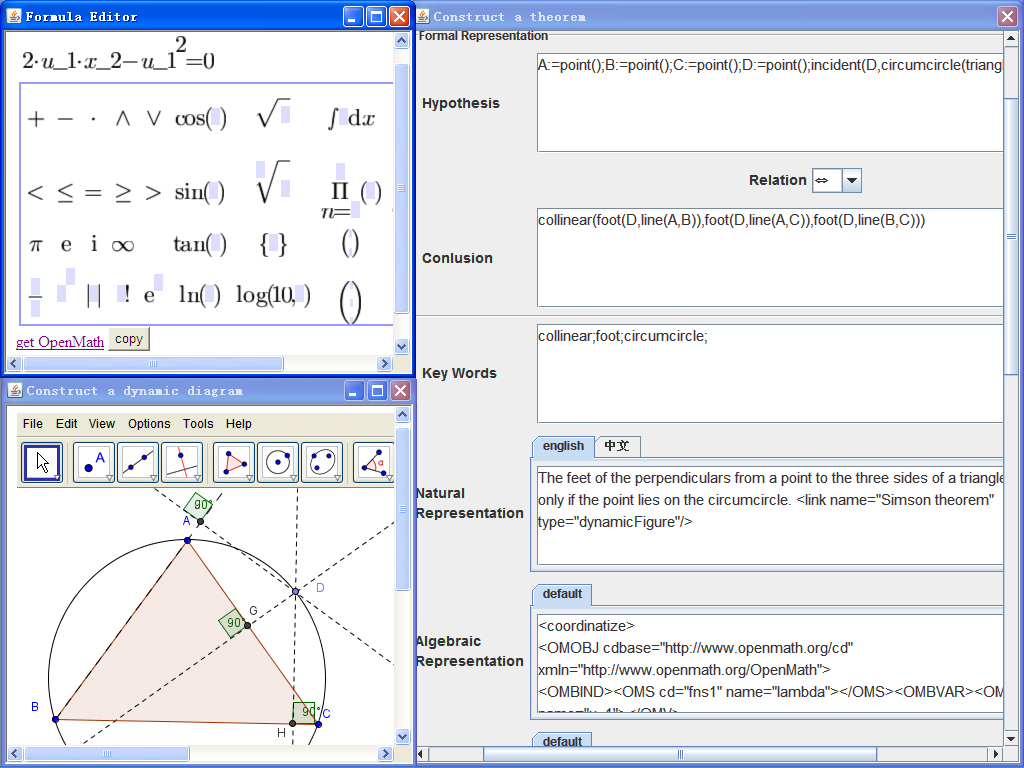}
\caption{Constructing \texttt{Simson's theorem}
object}\label{constructSimsontheorem}
\end{center}
\vspace{-0.5cm}
\end{figure}

The textbook knowledge base stores the knowledge data of knowledge
objects and category objects, as well as their relations. These
objects are the units that may be managed, retrieved, and processed
by the other modules. In order to identify and distinguish them, the
system automatically assigns each object a unique \emph{objectID}.
Then relational tables are defined that specify how data items are
related with the objects and what the relations among these objects
are. \cite{knowledgebase} We have created a database containing
these tables in MS SQL Server and chosen Java as the programming
language to develop the interfaces for users to maintain the
knowledge data of the Concept (Definition), Axiom, Lemma, Theorem,
Corollary, Conjecture, Problem, Example, Exercise, Proof, Solution,
Introduction, Remark, and Category objects as well as their
relations. Our system employs several external packages for editing
specific data. The dynamic mathematics software GeoGebra is used for
producing dynamic figures. The MathDox formula editor \cite{editor}
is used to create expressions encoded in OpenMath for the algebraic
representations. One can construct, for example, \texttt{Simson's
theorem} as in Fig.~\ref{constructSimsontheorem}.

Currently, the system provides simple query services for users who
may input search commands and view the results using keywords and
relations. The queries through relations work at the level of
knowledge objects and category objects. This means that the queries
need to be described by using the \emph{objectIDs} of the objects
but not simple natural texts. We explain the commands for queries
below.

\begin{itemize}

\item \texttt{keyWords[$word_1,\ldots,word_n$]} returns the set of knowledge
objects and category objects with keywords \texttt{$word_1$} and
$\ldots$ and \texttt{$word_n$}.

\item \texttt{relation[*,\,$objectID$,\,$relationType$]} returns the set of knowledge
objects that are each in the relation of
\emph{relationType} to the knowledge object
identified by \emph{objectID};
\item \texttt{relation[$objectID$,\,*,\,$relationType$]} returns the set
of knowledge objects such that the knowledge
object identified by \emph{objectID} is in the relation of
\emph{relationType} with each of them.

\end{itemize}

\begin{figure}
\begin{center}
\includegraphics[width=11cm]{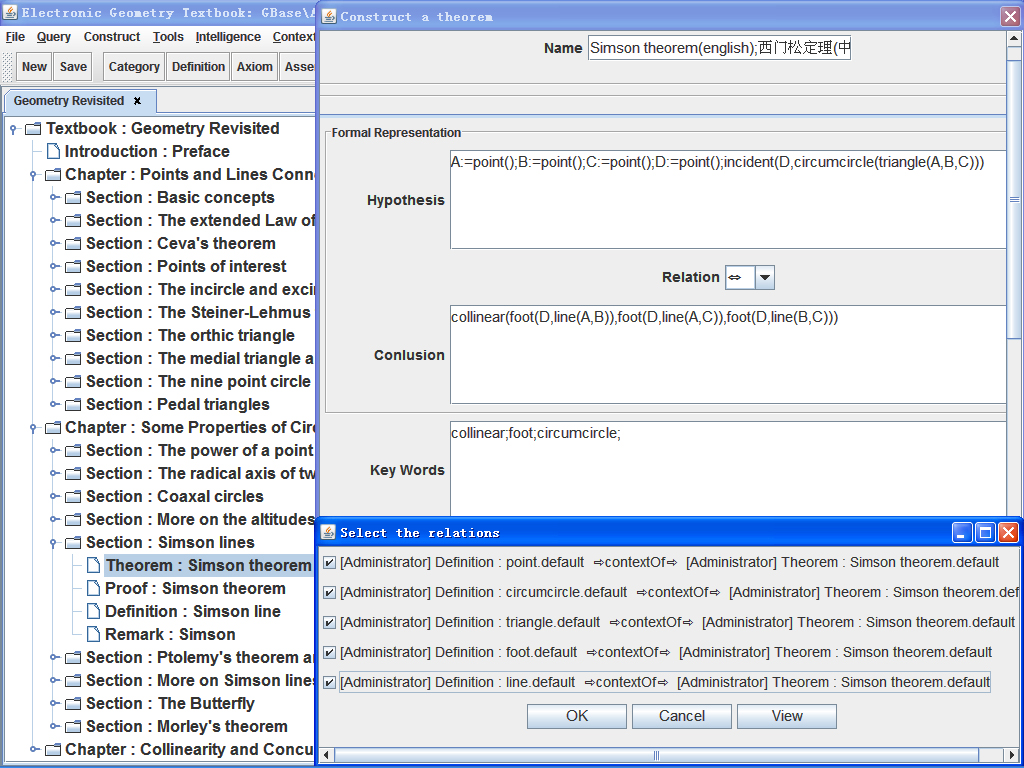}
\caption{Discovering the Context relations
automatically}\label{relationDiscovering}
\end{center}
\vspace{-0.5cm}
\end{figure}

The relations among knowledge objects and category objects are very
important for structuring the knowledge base. One way to acquire the
relations is manual annotation through reference to the
\emph{objectIDs} of the corresponding objects. We have implemented
another way that the system can automatically discover the Context
and Inheritance relations by matching the concept declarations with
the instances used in the formal representations of knowledge
objects. For example, in the process of constructing \texttt{Simson's
theorem}, the definitions of \texttt{point}, \texttt{line},
\texttt{foot}, \texttt{triangle}, \texttt{circumcircle} are
automatically found that provide the context for \texttt{Simson's
theorem}. The system will list the relations discovered for the user
to select. (see Fig.~\ref{relationDiscovering})

\subsection{User Interface of Electronic Geometry Textbook}

\begin{figure}
\begin{center}
\includegraphics[width=11cm]{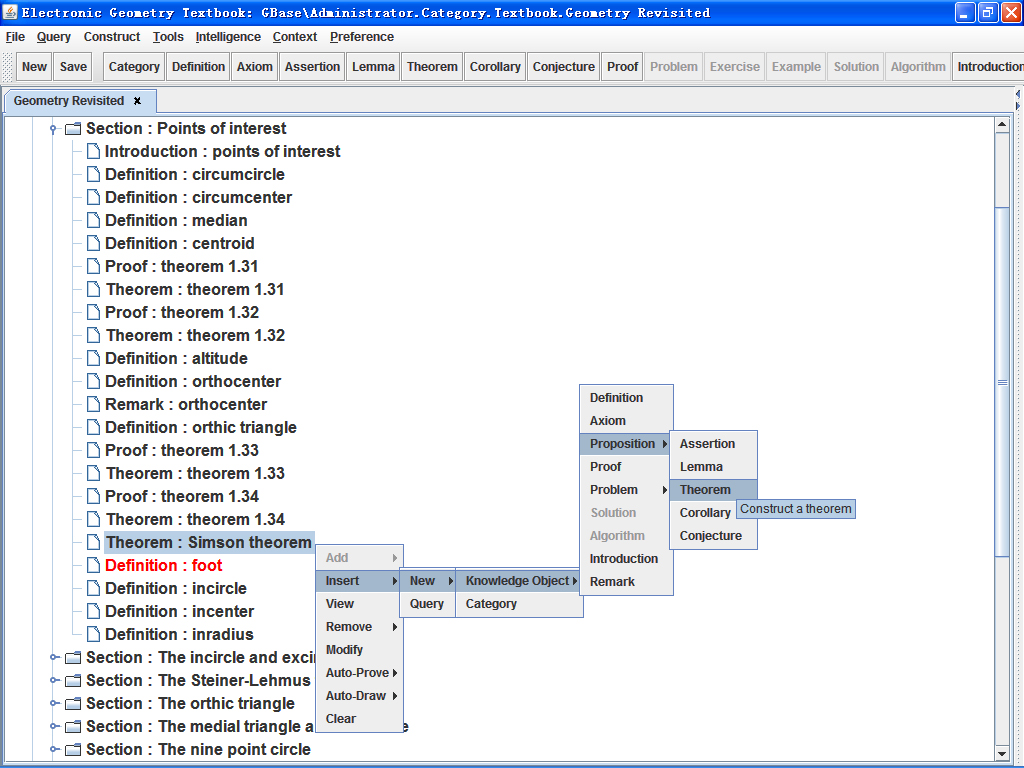}
\caption{The definition of \texttt{foot} is highlighted when
\texttt{Simson's theorem} is placed before
it.}\label{consistencyChecking}
\end{center}
\vspace{-0.5cm}
\end{figure}

With the textbook knowledge base created, a user interface is
implemented for users to construct dynamic textbooks, which are
rendered as trees. The category objects are rendered as branch nodes
and knowledge objects are rendered as tree leaves. Via dialogs, one
can construct textbooks interactively by adding, inserting,
removing, modifying, and rearranging the category objects and
knowledge objects, and annotating their relations one by one. These
objects may be newly created in, or fetched from, the knowledge
base. While performing these manipulations, the system can check
automatically whether the structure of the current textbook is
consistent. The user may be given tips if it is inconsistent and the
textbook should be restructured until it becomes consistent. For
example, if one places \texttt{Simson's theorem} before the
definition of \texttt{foot} (which provides the context for
\texttt{Simson's theorem}), then the system will highlight the node
of \texttt{foot}. (see Fig.~\ref{consistencyChecking})

\subsection{Presentation of Geometric Textbook Knowledge}

\begin{figure}
\begin{center}
\includegraphics[width=11cm]{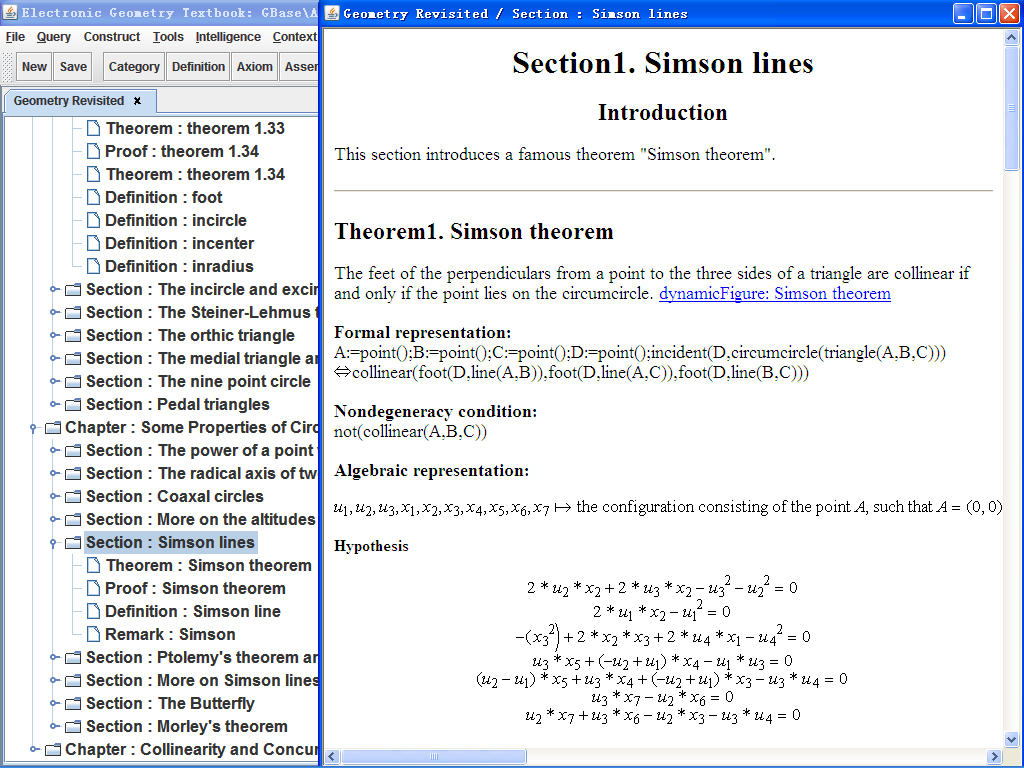}
\caption{Rendering the Section ``Simson lines" in
English}\label{browseEnglish}
\end{center}
\vspace{-0.5cm}
\end{figure}

\begin{figure}
\begin{center}
\includegraphics[width=11cm]{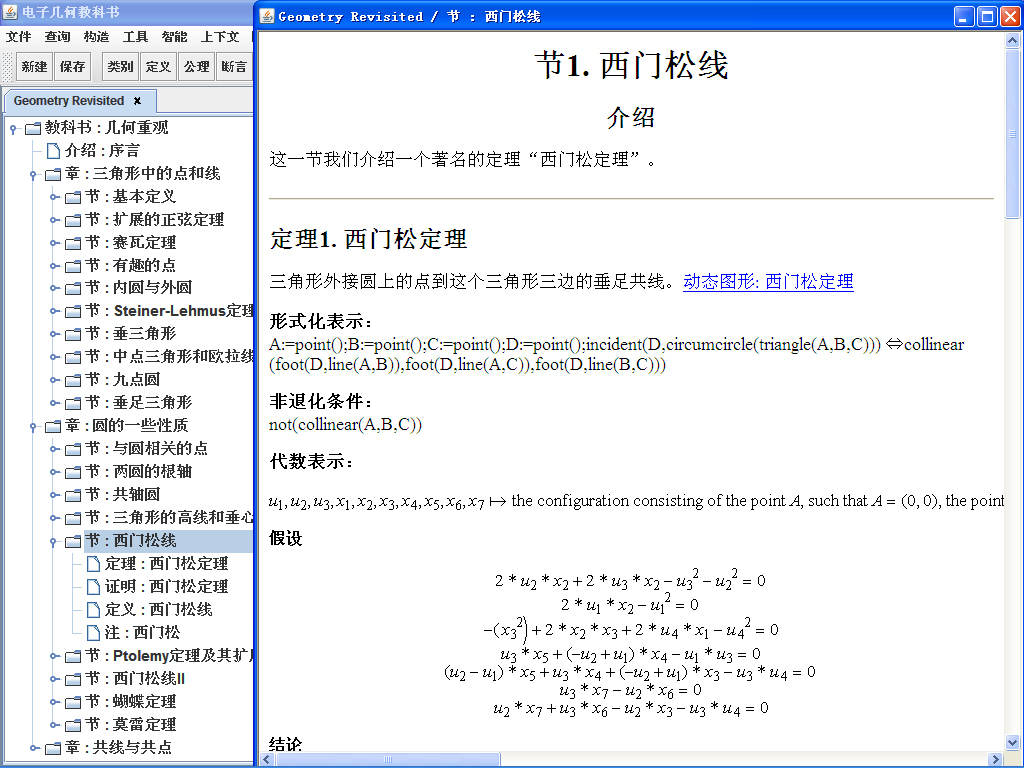}
\caption{Rendering the Section ``Simson lines" in
Chinese}\label{browseChinese}
\end{center}
\vspace{-0.5cm}
\end{figure}

For a textbook once constructed, it is necessary to provide a view that
presents the knowledge objects and category objects in readable
styles. From the textbook (or part of one), the system automatically
generates corresponding XML documents by assembling the data of the
selected objects and renders them (both in English and Chinese) via
JDesktop Integration Components (JDIC \cite{jdic}), which provide
Java applications with access to functionalities and facilities
furnished by the native desktop (see Figs.~\ref{browseEnglish},
\ref{browseChinese}). The generated XML documents can easily be
styled and transformed into other document formats (MathDox
\cite{mathdox}, or PDF, etc.) by using XSLT.

\subsection{Automatic Problem Solving}
\label{communicationsection}

\begin{figure}
\begin{center}
\includegraphics[width=12cm]{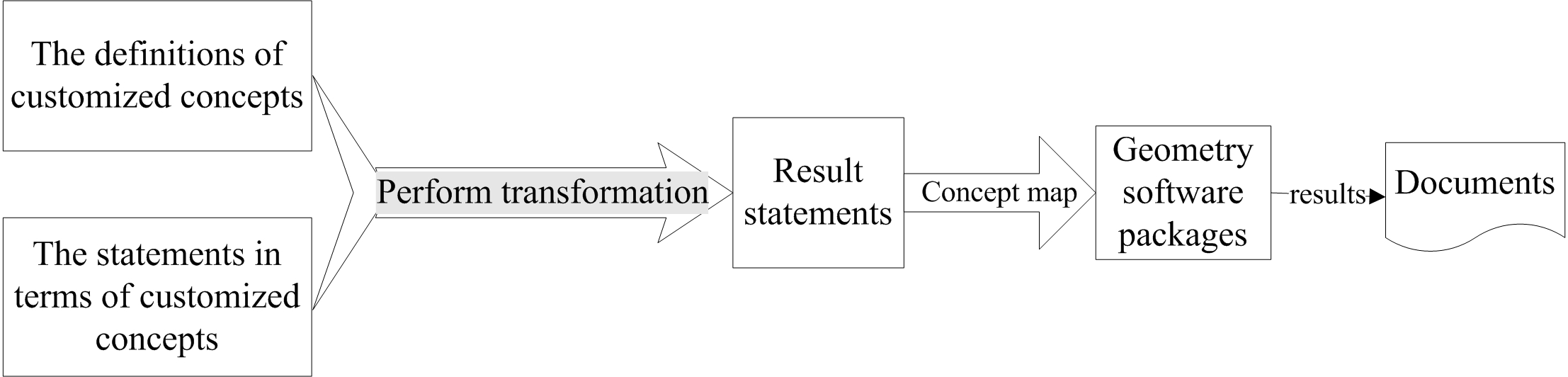}
\caption[1]{The workflow of communication with geometry software
packages}\label{communication}
\end{center}
\vspace{-0.5cm}
\end{figure}

As presented in Section~\ref{language}, geometric statements of
knowledge objects are formalized and represented by using customized
geometric concepts. However, most geometry software tools only
implement some of them. For communicating with the available
stand-alone packages, it is indispensable to transform these
statements into semantically equivalent ones employing the concepts
that the target geometry software packages are able to identify and
manipulate. Inspired by the idea of expression simplification in
functional programming, we have implemented this transformation
automatically by applying definitions of customized geometric
concepts stored in the knowledge base. The process of communication
with geometry software packages is diagrammed in
Fig.~\ref{communication}.

\begin{figure}
\begin{center}
\includegraphics[width=11cm]{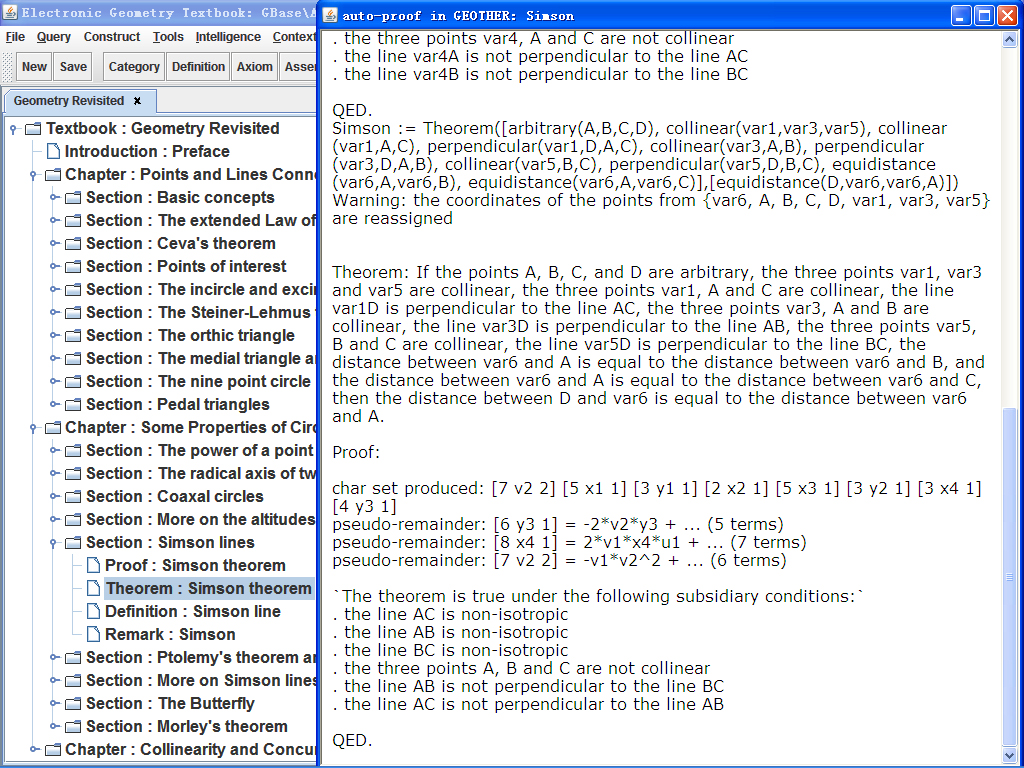}
\caption[1]{\texttt{Simson's theorem} in the textbook is automatically
proved by using GEOTHER.}\label{prove}
\end{center}
\vspace{-0.5cm}
\end{figure}

\begin{figure}
\begin{center}
\includegraphics[width=11cm]{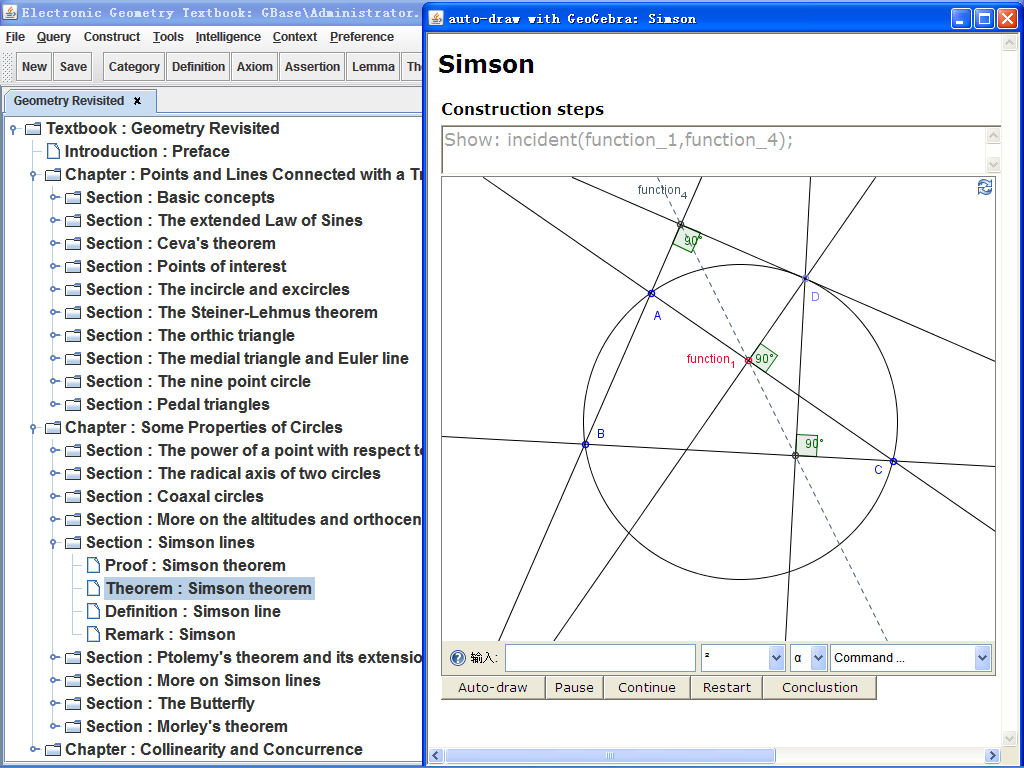}
\caption[1]{The dynamic figure of \texttt{Simson's theorem} is
automatically drawn by using GeoGebra.}\label{draw}
\end{center}
\vspace{-0.5cm}
\end{figure}

We have implemented communication with GEOTHER for automated
theorem proving (see Fig.~\ref{prove}) and with GeoGebra for drawing
dynamic figures automatically (see Fig.~\ref{draw}).

\section{Conclusion and Future Work}

This paper describes the design principles and, briefly, the
technical realization of the first version of Electronic Geometry
Textbook. The system provides an integrated environment for users to
manage and share textbook knowledge objects, construct dynamic
geometry textbooks interactively and efficiently, and publish styled
geometric documents easily. The knowledge objects encapsulate
multiple forms of knowledge data for different applications, such as
presentation in natural languages, processing by selected external
geometry software packages for automated theorem proving and dynamic
figure drawing, etc. The textbooks constructed can be manipulated
easily with automatic consistency-checking in real time. The system
can be viewed as a geometry-textbook-authoring assistant.

Currently, the development of Electronic Geometry Textbook is still
at an early stage. It is far from its ultimate goal of seeing that
the dynamic geometry textbooks constructed can be used in practice
with students.  For instance, communications with geometry software
packages lack interactions with users.  Aside from a series of
experiments on the system in the near future, we are preparing to
explore approaches to the design and development of interactive
exercises in geometry and to enhance the usability of the textbooks.

\bigskip\noindent {\bf Acknowledgments.} The author wishes to thank
Professor Dongming Wang for his kind supervision and Professor Arjeh
M.\ Cohen for inviting him to visit and work at the DAM group of
Technische Univ\-er\-siteit Eindhoven. His work has benefited from
communication with Hans Cuypers and other members of the DAM group
during his stay in Eindhoven. The author also wishes to thank the
referees for their insightful comments and Patrick Ion for the
copy-editing which have helped bring this paper to its present form.
It has been supported by the Chinese National Key Basic Research
(973) Project 2005CB321901/2 and the SKLSDE Open Fund
BUAA-SKLSDE-09KF-01.

\bibliography{bibfile}

\end{document}